\documentclass[11pt]{article}

\usepackage[final]{acl}
\usepackage{algorithm}
\usepackage{algpseudocode}
\usepackage{times}
\usepackage{latexsym}
\usepackage{amsmath}
\usepackage{amssymb}
\usepackage{xcolor}
\usepackage{booktabs}
\usepackage{enumitem}
\usepackage{placeins}

\definecolor{stepblue}{RGB}{80,150,220}
\newcommand{\AlgStep}[1]{\State \textcolor{stepblue}{$\blacktriangleright$ \textbf{#1}}}
\usepackage[T1]{fontenc}

\usepackage[utf8]{inputenc}

\usepackage{microtype}

\usepackage{inconsolata}

\usepackage{graphicx}

\title{CAMFT: Conflict-Aware Mergeable Fine-Tuning \\
for Large Language Models}

\author{
Jingang Zhou$^{1,2}$ \quad
Haiyang Guo$^{1,2}$ \quad
Yuan Ma$^{2}$ \quad
Han Zhu$^{2}$ \quad
Xu-Yao Zhang$^{1,2,*}$ \\
$^{1}$School of Advanced Interdisciplinary Sciences, University of Chinese Academy of Sciences \\
$^{2}$Institute of Automation, Chinese Academy of Sciences \\
\texttt{zhoujingang2025@ia.ac.cn} \quad \texttt{xyz@nlpr.ia.ac.cn} \\
}

\hypersetup{
  pdftitle={CAMFT: Conflict-Aware Mergeable Fine-Tuning for Large Language Models},
  pdfauthor={Jingang Zhou, Haiyang Guo, Yuan Ma, Han Zhu, Xu-Yao Zhang}
}

\begin{document}
\maketitle
\begingroup
\renewcommand{\thefootnote}{\fnsymbol{footnote}}
\footnotetext[1]{Corresponding author.}
\endgroup

\begin{abstract}
Model merging has emerged as a promising paradigm for integrating multiple task-specific capabilities into a single large language model. However, existing methods predominantly focus on post-hoc processing of independently fine-tuned models, overlooking how the training phase itself impacts cross-task compatibility. Resolving parameter conflicts after fine-tuning is inherently sub-optimal. To address this, we propose \textbf{CAMFT}, a \textbf{C}onflict-\textbf{A}ware \textbf{M}ergeable \textbf{F}ine-\textbf{T}uning method that makes task adaptation both efficient and merge-aware.
CAMFT treats mergeability as a property shaped during fine-tuning, rather than only a problem to be solved after fine-tuning.
By guiding each task to update sparse coordinates with lower cross-task conflict, CAMFT produces task updates that are efficient to train and more compatible for downstream model merging.
Extensive experiments demonstrate that CAMFT outperforms standard fine-tuning baselines in multi-task merging scenarios. Codes are available at \url{https://github.com/gyanchow/CAMFT-LLM}.
\end{abstract}

\section{Introduction}
\label{sec:introduction}

Large language models (LLMs) have become general-purpose backbones for a wide range of downstream tasks~\citep{10.5555/3495724.3495883, JMLR:v24:22-1144}.
As these models are adapted to increasingly diverse domains, a central challenge is how to integrate multiple task-specific capabilities into a single model without repeatedly training and deploying separate models.
A straightforward solution is multi-task fine-tuning, which jointly trains one model on the union of all task data ~\citep{sanh2022multitask, wei2022finetuned}.
However, this paradigm is often impractical in practice: task data may be distributed across different users, organizations, or time periods; new tasks may arrive after initial training; and retraining large models on accumulated data can be prohibitively expensive.
These limitations motivate modular adaptation, where task-specific knowledge is learned independently and later integrated into a unified model.

\begin{figure}[t]
    \centering
    \includegraphics[width=\columnwidth]{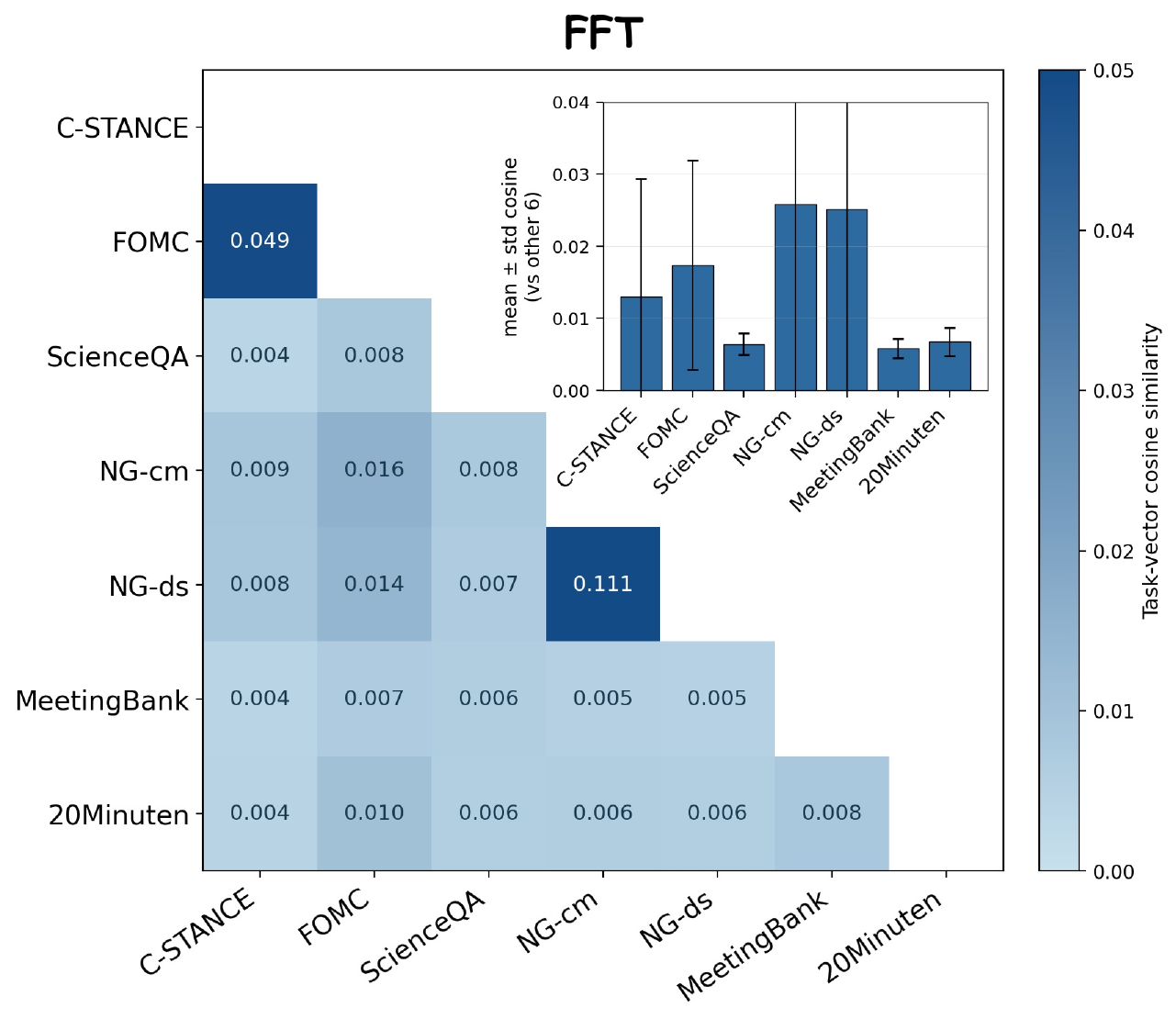}
\caption{
    Pairwise cosine similarities among task vectors obtained by independently full-parameter fine-tuning on seven tasks.
    }
    \label{fig:intro_fft_conflict}
\end{figure}

\begin{figure*}[t]
    \centering
    \includegraphics[width=0.96\textwidth]{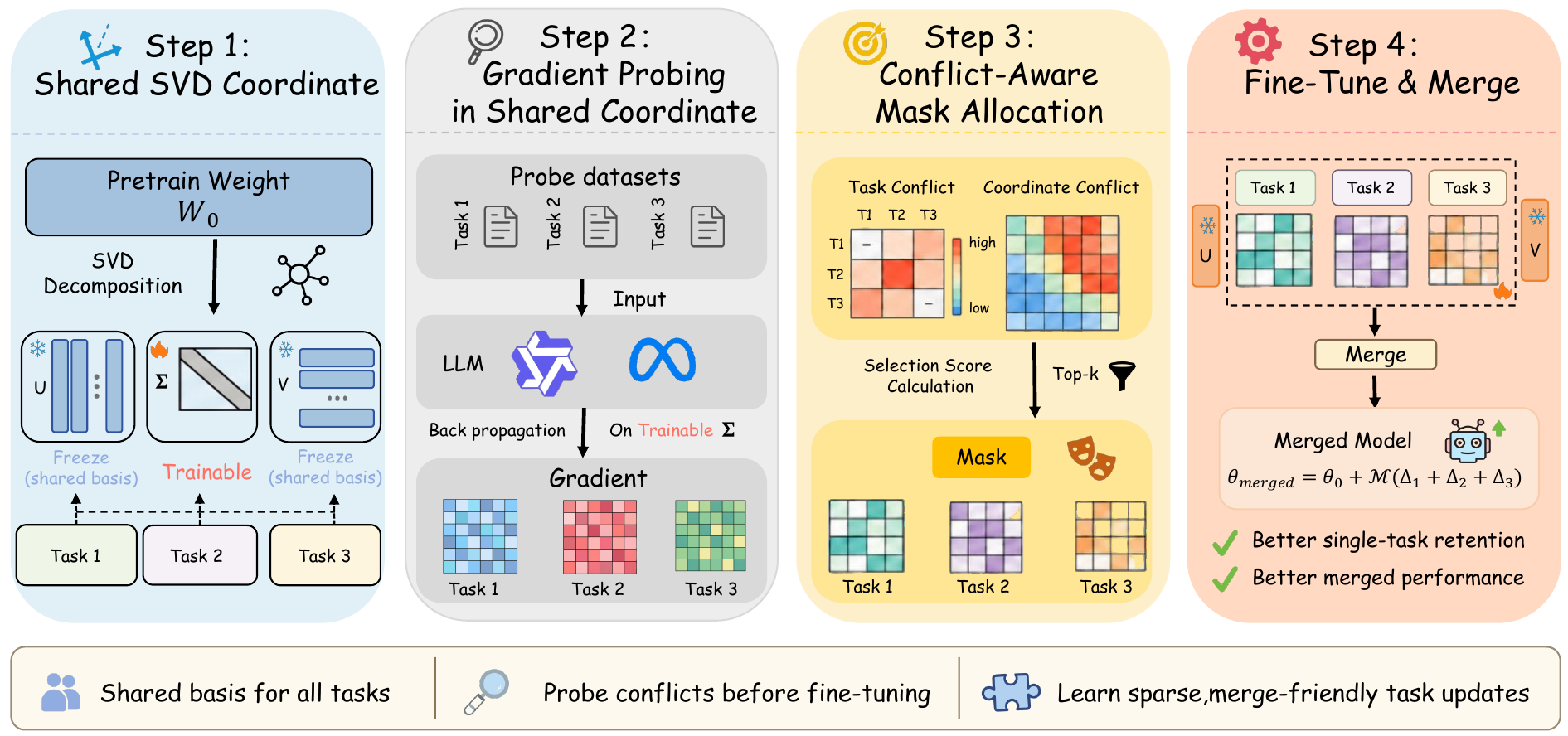}
    \caption{
    Overview of CAMFT.
    CAMFT constructs a shared SVD coordinate system and uses gradient probing to quantify cross-task conflicts. It then allocates sparse, conflict-aware masks to each task. By isolating parameter updates, CAMFT ensures the independently fine-tuned models are highly compatible for merging.
    }
    \label{fig:main}
\end{figure*}

One prominent solution addresses this problem through task vectors~\citep{ilharco2023editing} and model merging~\citep{yang2026model}, which arithmetically combine parameter updates to enable data-free, modular reuse of task-specific expertise. Recent merging methods improve fusion quality through post-hoc techniques such as importance weighting, sign alignment, and sparse resampling~\citep{yadav2023tiesmerging, pmlr-v235-yu24p, matena2022merging, yang2024adamerging, zeng2026robustmerge}. These underlying task updates are typically obtained via full-parameter fine-tuning or parameter-efficient fine-tuning (PEFT) methods like LoRA~\citep{pmlr-v97-houlsby19a, hu2022lora} to reduce training costs. However, this entire paradigm operates strictly at the \emph{post-hoc fusion} stage. Regardless of whether tasks update all parameters or an efficient subspace, independent fine-tuning inevitably writes incompatible information into overlapping parameter directions. Figure~\ref{fig:intro_fft_conflict} shows that task vectors obtained by independent full-parameter fine-tuning are not mutually orthogonal, indicating substantial parameter-space entanglement before merging. By attempting to resolve these interferences strictly after the fact, both post-hoc merging and parameter-efficient methods overlook the opportunity to optimize cross-task compatibility during training.

Recent work has begun to explore merge-aware fine-tuning by explicitly considering mergeability during training~\citep{lee2025mitigating, yang2026mergopt, zhang-zhou-2025-unraveling}. While these methods improve fusion outcomes through strategies such as sharpness minimization or subspace orthogonalization, they typically operate at a coarse granularity. By enforcing global properties or subspace-level separation, they overlook a fundamental question: \emph{which specific parameter coordinates should each task update to minimize merge conflict?} Answering this question is crucial, because even if tasks are separated in broad subspaces, updating overlapping parameter coordinates will still cause severe weight-level interference during merging.

To answer this question, we argue that effective modular adaptation must deliberate not only \emph{how many} parameters a task updates, but specifically \emph{where} it is allowed to place them.
If multiple tasks update the same parameter directions with inconsistent requirements, post-hoc merging inevitably compresses these incompatible changes, leading to performance collapse.
Instead of treating this conflict as noise to be resolved after training, we leverage it as a constructive signal to shape the updates beforehand.
Motivated by this insight, we propose \textbf{CAMFT} (\textbf{C}onflict-\textbf{A}ware \textbf{M}ergeable \textbf{F}ine-\textbf{T}uning).
As illustrated in Figure~\ref{fig:main}, CAMFT constructs a shared SVD coordinate system from the pretrained weights to explicitly probe cross-task conflicts prior to fine-tuning.
By quantifying these conflicts, CAMFT allocates sparse trainable coordinates for each task, guiding their adaptations toward mutually compatible regions.
Consequently, CAMFT bridges PEFT and model merging by producing task updates that are intrinsically merge-friendly.
Extensive evaluations on the diverse TRACE benchmark~\citep{trace2023} demonstrate that this explicit coordinate allocation significantly improves the fused model's performance over standard fine-tuning while fully preserving competitive single-task capabilities.

Our contributions are summarized as follows:
\begin{itemize}
    \item We introduce a novel perspective for merge-aware fine-tuning by highlighting the need for weight-level interference control, emphasizing \emph{where} a task should update its parameters.

    \item We formulate multi-task adaptation as a conflict-aware coordinate allocation problem and propose CAMFT. By leveraging gradient probing within a shared spectral space, CAMFT explicitly guides task updates toward less conflicting coordinates.

    \item Experiments on the TRACE benchmark demonstrate that CAMFT consistently improves merged performance across various downstream merging algorithms while strictly preserving single-task capabilities.
\end{itemize}

\section{Related Work}
\label{sec:related_work}

\paragraph{Spectral PEFT.}
PEFT methods adapt pretrained models with a small number of trainable parameters, often through structured low-rank updates such as LoRA~\citep{hu2022lora}. Recent spectral PEFT methods use the SVD of pretrained weights to define better adaptation subspaces: PiSSA~\citep{meng2024pissa} updates principal singular components, MiLoRA~\citep{wang-etal-2025-milora} adapts minor singular components, and SVFT~\citep{NEURIPS2024_48c368f1} learns sparse coefficients over singular-vector directions. Unlike these methods, which mainly improve single-task adaptation, CAMFT uses the SVD basis to compare task updates and prevent cross-task conflicts during fine-tuning.
\paragraph{Model merging.}
Post-hoc merging methods combine task-specific models after fine-tuning. Weight averaging and Task Arithmetic directly merge model parameters or task vectors~\citep{pmlr-v162-wortsman22a, ilharco2023editing}, but can suffer from conflicts among task updates. Fisher merging and AdaMerging estimate merging weights from parameter importance or unlabeled data~\citep{matena2022merging, yang2024adamerging}, while TIES and DARE reduce conflicts by pruning, sign alignment, or rescaling~\citep{yadav2023tiesmerging, pmlr-v235-yu24p}. In contrast, CAMFT aims to prevent conflicts during fine-tuning and remains compatible with post-hoc merging algorithms.

\paragraph{Merge-aware fine-tuning.}
Recent work shows that fine-tuning strategies can substantially affect model mergeability. SAFT~\citep{lee2025mitigating} uses sharpness-aware optimization to find flatter task-specific minima and reduce parameter interference. MergOPT~\citep{yang2026mergopt} improves merging robustness by simulating merge-induced perturbations during training. OSRM~\citep{zhang-zhou-2025-unraveling} constrains LoRA updates into orthogonal subspaces to mitigate cross-task interference before merging. These methods mainly enhance global robustness or enforce subspace-level separation, whereas CAMFT explicitly identifies coordinate-level conflicts in a shared spectral basis and allocates trainable regions accordingly.

\section{Method}
\label{sec:method}

CAMFT addresses model merging at the fine-tuning stage by constraining \emph{where} each task is allowed to update its parameters in a shared spectral basis, rather than resolving conflicts post hoc.
We formulate merge-aware fine-tuning as a sparse coordinate allocation problem and present the method in five parts.
Section~\ref{sec:problem_setup} formalizes the multi-task fine-tuning and merging setup.
Section~\ref{sec:svd_coordinate} introduces a shared SVD coordinate system from the pretrained weights, in which task updates from different tasks can be compared coordinate by coordinate.
Section~\ref{sec:conflict_probing} estimates per-coordinate importance and cross-task conflict from a small probing set.
Section~\ref{sec:mask_allocation} allocates a sparse trainable mask for each task that balances task importance against shared-coordinate conflict.
Section~\ref{sec:training_merging} describes masked fine-tuning of each task and the post-hoc merging step.

\subsection{Problem Setup}
\label{sec:problem_setup}

Let $\theta_0$ denote the pretrained parameters and $\mathcal{T}=\{1,\dots,T\}$ the set of tasks.
Fine-tuning task $t$ produces an update $\Delta_t$, yielding $\theta_t=\theta_0+\Delta_t$.
A merger $\mathcal{M}$ combines these updates as
\begin{equation}
\begin{aligned}
    \Delta_{\mathrm{merge}}
    &= \mathcal{M}(\Delta_1,\dots,\Delta_T), \\
    \theta_{\mathrm{merge}}
    &= \theta_0+\Delta_{\mathrm{merge}}
\end{aligned}
\end{equation}
CAMFT does not alter $\mathcal{M}$; it changes how each $\Delta_t$ is produced so that the resulting updates are more compatible before merging.
\subsection{Shared SVD Coordinate System}
\label{sec:svd_coordinate}

To compare and control task updates across different tasks, CAMFT first establishes a shared coordinate system for each target linear layer.
Inspired by \textbf{SVFT}~\citep{NEURIPS2024_48c368f1}, we choose the singular value decomposition (SVD) of pretrained weights as this shared basis because it provides a natural, task-agnostic decomposition of the parameter space.

For a pretrained weight matrix $W_0\in\mathbb{R}^{d_{\mathrm{out}}\times d_{\mathrm{in}}}$, we compute its compact full-rank SVD:
\begin{equation}
    W_0 = U\Sigma V^\top, \quad r=\min(d_{\mathrm{out}},d_{\mathrm{in}}).
\end{equation}
where $U \in \mathbb{R}^{d_{\mathrm{out}}\times r}$ and $V \in \mathbb{R}^{d_{\mathrm{in}}\times r}$ are orthonormal matrices containing the left and right singular vectors, and $\Sigma \in \mathbb{R}^{r\times r}$ is a diagonal matrix of singular values.
The pretrained weight $W_0$ and its SVD bases $U$ and $V$ are frozen throughout training.

For each task $t$, CAMFT learns a residual update in this SVD coordinate system.
We introduce a trainable coordinate matrix $M_t \in \mathbb{R}^{r\times r}$ and a binary mask $B_t \in \{0,1\}^{r\times r}$.
The effective update is:
\begin{equation}
    \Delta W_t = \alpha \cdot U (M_t \odot B_t) V^\top,
    \label{eq:svd_update}
\end{equation}
where $\odot$ denotes element-wise multiplication and $\alpha$ is a scaling factor.
We denote each entry of $M_t$ by its index $p=(i,j)$, which represents the interaction between the $i$-th left singular direction and the $j$-th right singular direction.
The set of all coordinates is denoted by $\Omega_\ell$, with $|\Omega_\ell| = r^2$.

This SVD-based parameterization places all task updates in a common coordinate system, so their locations can be compared by coordinate.
The singular directions of the pretrained weights define a model-specific basis for allocating update capacity, and the mask $B_t$ determines which coordinates task $t$ is allowed to modify.
CAMFT therefore formulates merge-aware fine-tuning as a sparse coordinate allocation problem, restricting each task update to coordinates that are important for the task and less likely to conflict with other tasks.

\subsection{Probing Coordinate Importance and Conflict}
\label{sec:conflict_probing}

Before fine-tuning, CAMFT probes task gradients to estimate two coordinate-level quantities: how strongly each task depends on a coordinate and whether different tasks prefer inconsistent update directions on that coordinate.

\noindent\textbf{Gradient probing.}
For each task $t$, we sample a small probing set $\mathcal{D}^{\mathrm{probe}}_t$ from the task's training data.
Starting from the pretrained model, we compute gradients with respect to the SVD-coordinate matrix $M_t$ at each coordinate $p \in \Omega_\ell$.
For each coordinate, CAMFT records two statistics from the gradient distribution:
\begin{equation}
\begin{aligned}
    I_{t,p}
    &= \mathbb{E}_{(x,y)\sim\mathcal{D}^{\mathrm{probe}}_t}
    \left[\left|\nabla_{M_p}\ell_t(x,y)\right|\right], \\
    R_{t,p}
    &= \mathbb{E}_{(x,y)\sim\mathcal{D}^{\mathrm{probe}}_t}
    \left[\nabla_{M_p}\ell_t(x,y)\right].
\end{aligned}
\label{eq:probe_scores}
\end{equation}
where $\ell_t(x,y)$ is the task-specific loss function.
The \emph{importance score} $I_{t,p} \geq 0$ measures how strongly coordinate $p$ influences task $t$'s loss.
The \emph{signed gradient} $R_{t,p} \in \mathbb{R}$ captures the preferred update direction for coordinate $p$.
Both statistics are essential: two tasks may have high importance on the same coordinate, but whether this causes merge interference depends on whether they prefer similar or opposite directions.

\noindent\textbf{Task specificity.}
Raw importance scores are not directly comparable across layers due to differing gradient scales.
We therefore apply layer-wise rank normalization to obtain $\widetilde I_{t,p} \in [0,1]$.
Using these normalized scores, we define the \emph{task specificity} of coordinate $p$ for task $t$:
\begin{equation}
    D_{t,p} = \frac{\widetilde I_{t,p}}{\sum_{j=1}^{T}\widetilde I_{j,p}+\epsilon}.
    \label{eq:specificity}
\end{equation}
The specificity $D_{t,p}$ represents the fraction of total demand for coordinate $p$ that comes from task $t$.
When $D_{t,p} \approx 1$, the coordinate is \emph{task-specific}: it is crucial for task $t$ but not other tasks.
When $D_{t,p}$ is more uniformly distributed across tasks, the coordinate is \emph{shared}: multiple tasks rely on it.

\noindent\textbf{Coordinate conflict score.}
A shared coordinate is not necessarily problematic for merging.
If all tasks prefer similar update directions, they can share the coordinate without interference.
The coordinate becomes risky only when it is both shared and subject to heterogeneous update directions.
To capture this intuition, we define a \emph{conflict score} that combines task overlap with directional heterogeneity.

First, we measure how broadly a coordinate is shared using the normalized entropy of the specificity distribution:
\begin{equation}
    O_p = -\frac{1}{\log T}\sum_{t=1}^{T} D_{t,p} \log(D_{t,p}+\epsilon),
    \label{eq:overlap}
\end{equation}
where $O_p \approx 0$ indicates a task-specific coordinate and $O_p \approx 1$ indicates a broadly shared one.

Second, we measure directional heterogeneity by normalizing signed gradients within each task and layer, then computing the variance across tasks:
\begin{equation}
\begin{aligned}
    \widehat R_{t,p}
    &=
    \frac{R_{t,p}}{s_{t,\ell}}, \\
    s_{t,\ell}
    &=
    \sqrt{
    \frac{1}{|\Omega_\ell|}
    \sum_{q\in\Omega_\ell} R_{t,q}^2
    + \epsilon } .
\end{aligned}
\label{eq:norm_signed}
\end{equation}
\begin{equation}
    C_p = O_p \cdot \mathrm{Var}_{t\in\mathcal{T}}\left(\widehat R_{t,p}\right).
    \label{eq:conflict_score}
\end{equation}
The conflict score $C_p$ is high only when both conditions hold: the coordinate is shared by multiple tasks ($O_p$ large), and the tasks prefer different update directions (variance large).
This design ensures that we penalize coordinates that genuinely cause interference, not simply those shared across tasks.

\subsection{Conflict-Aware Mask Allocation}
\label{sec:mask_allocation}

Based on the probing statistics, CAMFT allocates a sparse trainable mask $B_t$ for each task $t$.
The mask determines which SVD coordinates the task is permitted to update during fine-tuning.
Our goal is to assign each task to coordinates that are both important for its performance and unlikely to conflict with other tasks.

\noindent\textbf{Selection score.}
For each task-coordinate pair $(t,p)$, we compute a \emph{selection score} that balances three considerations:
\begin{equation}
\begin{aligned}
    S_{t,p}
    &=
    \frac{
    \widetilde I_{t,p}(1+\gamma D_{t,p})
    }{
    1+\lambda C_p(1-D_{t,p})
    } .
\end{aligned}
\label{eq:selection_score}
\end{equation}
\begin{itemize}[leftmargin=*,nosep]
    \item \textbf{Importance} ($\widetilde I_{t,p}$): Prioritizes coordinates that the task relies on.
    \item \textbf{Specificity} ($D_{t,p}$): Rewards coordinates specific to the current task, reducing competition. The hyperparameter $\gamma \geq 0$ controls this reward.
    \item \textbf{Conflict} ($C_p$): Penalizes shared high-conflict coordinates. The penalty is attenuated when the coordinate is specific to the current task ($D_{t,p} \approx 1$), ensuring that task-critical coordinates remain available even if they occasionally conflict. The hyperparameter $\lambda \geq 0$ controls the conflict penalty strength.
\end{itemize}

\noindent\textbf{Top-$k$ mask construction.}
For each layer $\ell$, we allocate a fixed budget of trainable coordinates.
Given a sparsity ratio $\rho \in (0,1]$, each task selects the top-$k$ coordinates according to $S_{t,p}$, where $k_\ell = \lfloor \rho \cdot |\Omega_\ell| \rfloor$.
We first define the selected coordinate set:
\begin{equation}
    \mathcal{K}_{t,\ell}
    =
    \mathrm{TopK}_{q \in \Omega_\ell}(S_{t,q}, k_\ell),
\end{equation}
and construct the binary mask as
\begin{equation}
    B_{t,p}
    =
    \mathbb{1}\left[p \in \mathcal{K}_{t,\ell}\right],
\end{equation}
where $\mathbb{1}[\cdot]$ is the indicator function.
Coordinates with $B_{t,p}=1$ are trainable; all others remain fixed at zero during fine-tuning.

\noindent\textbf{High-conflict exclusivity.}
The selection score in Eq.~\eqref{eq:selection_score} discourages high-conflict coordinates through a soft penalty, but highly important coordinates may still be selected by multiple tasks.
To further reduce severe interference, CAMFT applies an \emph{exclusivity rule} to the most conflict-prone coordinates.

Let $\mathcal{H}_\ell$ denote the set of top-$h\%$ coordinates with the highest conflict scores $C_p$.
For each coordinate $p \in \mathcal{H}_\ell$, only the task with the highest specificity is allowed to retain it:
\begin{equation}
    t^*(p) = \arg\max_{t \in \mathcal{T}} D_{t,p}.
\end{equation}
For all other tasks $t \neq t^*(p)$, coordinate $p$ is removed from their masks: $B_{t,p} \leftarrow 0$.
If removal causes a task's mask to have fewer than $k_\ell$ trainable coordinates, we refill with the next highest-scoring available coordinates (excluding those in $\mathcal{H}_\ell$ assigned to other tasks), preserving the per-task budget.

Together, the selection score and exclusivity rule assign coordinates according to both task specificity and cross-task compatibility.
This allows CAMFT to preserve task-critical updates while restricting shared high-conflict coordinates that are likely to cause destructive interference during merging.

\begin{algorithm}[!t]
\small
\caption{Conflict-Aware Fine-Tuning}
\label{alg:camft}
\begin{algorithmic}[1]
\Require Pretrained model $\theta_0$; tasks $\mathcal{T}$; budget $\rho$.
\Ensure Merged model $\theta_{\mathrm{merge}}$.

\AlgStep{Step 1: SVD coordinate construction}
\For{each target layer $W_0$}
    \State Decompose $W_0=U\Sigma V^\top$.
    \State Freeze $U,V$.
\EndFor

\AlgStep{Step 2: Coordinate-level conflict estimation}
\For{each task $t\in\mathcal{T}$}
    \State Probe gradients on task data to estimate coordinate importance $I_{t,p}$ and signed update direction $R_{t,p}$.
\EndFor
\State Normalize importance scores $\{I_{t,p}\}$ to obtain $\{\widetilde{I}_{t,p}\}$.
\State Compute task specificity $D_{t,p}$ and coordinate conflict $C_p$.

\AlgStep{Step 3: Conflict-aware mask allocation}
\For{each task $t\in\mathcal{T}$}
    \State Compute selection score $S_{t,p}$ using $\widetilde{I}_{t,p}$, $D_{t,p}$, and $C_p$.
    \State Select top-$\rho$ coordinates in each layer to construct task mask $B_t$.
\EndFor
\State Apply high-conflict exclusivity and refill masks to keep the same budget.

\AlgStep{Step 4: Masked fine-tuning and merging}
\For{each task $t\in\mathcal{T}$}
    \State Fine-tune masked SVD update:
    \Statex \hspace{\algorithmicindent}
    $\Delta W_t=\alpha U(M_t\odot B_t)V^\top$.
    \State Reconstruct and save task update $\Delta_t$.
\EndFor
\State Merge task updates:
\Statex \hspace{\algorithmicindent}
$\Delta_{\mathrm{merge}}=\mathcal{M}(\Delta_1,\dots,\Delta_T)$.
\State \Return $\theta_0+\Delta_{\mathrm{merge}}$.

\end{algorithmic}
\end{algorithm}

\subsection{Task Fine-Tuning and Merging}
\label{sec:training_merging}

\begin{table*}[!ht]
\centering
\small
\setlength{\tabcolsep}{3.5pt}
\renewcommand{\arraystretch}{1.10}
\begin{tabular}{l c c c c c c c c c c}
\toprule
& & \multicolumn{7}{c}{\textbf{Single-task Performance}} & & \\
\cmidrule(lr){3-9}
\textbf{Method} & \textbf{Params} & \textbf{C-ST} & \textbf{FOMC} & \textbf{SciQA} & \textbf{NG-cm} & \textbf{NG-ds} & \textbf{MtgBk} & \textbf{20Min} & \shortstack{\textbf{Avg}\\\textbf{($n=3$)}} & \textbf{Merge.} \\
\midrule
Full FT & Full & 0.500 & 0.649 & 0.791 & 0.407 & 0.588 & 0.649 & \textbf{0.400} & \textbf{0.569$\pm$0.002} & -- \\
LoRA & 45.1M & 0.451 & \textbf{0.679} & \textbf{0.848} & 0.247 & 0.619 & 0.664 & 0.399 & 0.558$\pm$0.004 & -- \\
PiSSA & 45.1M & 0.431 & 0.653 & 0.835 & 0.264 & 0.603 & \textbf{0.673} & 0.382 & 0.549$\pm$0.002 & -- \\
SVFT & 34.4M & 0.423 & 0.626 & 0.617 & 0.257 & 0.533 & 0.526 & 0.365 & 0.478$\pm$0.008 & -- \\
OSRM$^\dagger$ & 45.1M & 0.436 & 0.673 & 0.845 & 0.231 & 0.566 & 0.643 & 0.384 & 0.540$\pm$0.005 & -- \\
MergOPT$^\dagger$ & Full & 0.495 & 0.673 & 0.678 & 0.413 & 0.582 & 0.651 & 0.392 & 0.555$\pm$0.008 & -- \\
CAMFT$^\dagger$ & 34.4M & \textbf{0.502} & 0.675 & 0.661 & \textbf{0.494} & \textbf{0.640} & 0.530 & \textbf{0.400} & 0.557$\pm$0.003 & -- \\
\midrule
& & \multicolumn{7}{c}{\textbf{Merged Performance}} & & \\
\cmidrule(lr){3-9}
\textbf{Method} & \textbf{Params} & \textbf{C-ST} & \textbf{FOMC} & \textbf{SciQA} & \textbf{NG-cm} & \textbf{NG-ds} & \textbf{MtgBk} & \textbf{20Min} & \shortstack{\textbf{Avg}\\\textbf{($n=3$)}} & \textbf{Merge.} \\
\midrule
Full FT & Full & 0.449 & 0.587 & \textbf{0.522} & 0.296 & 0.443 & 0.241 & 0.389 & 0.418$\pm$0.006 & 73.5\% \\
LoRA & 45.1M & 0.446 & 0.613 & 0.173 & 0.099 & 0.557 & 0.020 & \textbf{0.394} & 0.329$\pm$0.007 & 59.0\% \\
PiSSA & 45.1M & 0.444 & 0.609 & 0.265 & 0.187 & 0.532 & 0.165 & 0.382 & 0.369$\pm$0.005 & 67.2\% \\
SVFT & 34.4M & 0.398 & 0.573 & 0.357 & 0.203 & 0.326 & 0.253 & 0.355 & 0.352$\pm$0.008 & 73.6\% \\
OSRM$^\dagger$ & 45.1M & 0.425 & \textbf{0.659} & 0.282 & \textbf{0.309} & 0.511 & 0.184 & 0.391 & 0.394$\pm$0.004 & 72.9\% \\
MergOPT$^\dagger$ & Full & 0.468 & 0.588 & 0.424 & 0.235 & \textbf{0.582} & 0.406 & 0.390 & 0.442$\pm$0.002 & 79.7\% \\
CAMFT$^\dagger$ & 34.4M & \textbf{0.470} & 0.599 & 0.485 & 0.259 & 0.572 & \textbf{0.474} & 0.391 & \textbf{0.464$\pm$0.002} & \textbf{83.3\%} \\
\bottomrule
\end{tabular}
\caption{
Performance comparison on TRACE 7-task benchmark with Llama-3.2-1B.
\textbf{Single}: single-task performance before merging.
\textbf{Merged}: performance after merging all 7 tasks with TIES.
Task columns show the original fixed-seed breakdown; \textbf{Avg} is the mean $\pm$ standard deviation of the macro-average over three random seeds; \textbf{Merge.} is the ratio of the corresponding three-seed means.
\textbf{Params}: trainable parameters per task (M = million; Full = full-model fine-tuning).
All task metrics are normalized to [0,1]; best results are \textbf{bolded}.
$^\dagger$: methods designed for mergeability.
}
\label{tab:main_results_llama}
\end{table*}

Once the masks $\{B_t\}_{t=1}^T$ are constructed, each task is fine-tuned independently from the same pretrained model $\theta_0$.
For task $t$, we optimize only the entries of $M_t$ where the mask is active:
\begin{equation}
\begin{aligned}
    \min_{M_t}\quad
    \mathbb{E}_{(x,y)\sim\mathcal{D}_t}
    \left[
    \ell_t\left(
    f_{\theta_0 + \Delta_t}(x), y
    \right)
    \right],
\end{aligned}
\end{equation}
where $\mathcal{D}_t$ is the task's training data, $\ell_t$ is the task-specific loss, and $\Delta_t$ is reconstructed via Eq.~\eqref{eq:svd_update}.
All pretrained weights and SVD bases ($W_0$, $U$, $\Sigma$, $V$) remain frozen throughout; only the masked entries of $M_t$ are updated.
This ensures that each task learns within its allocated coordinate region.

After training, we reconstruct the task updates $\{\Delta_t\}_{t=1}^T$ into the original weight space.
These reconstructed updates can then be merged using any standard model merging algorithm:
\begin{equation}
\begin{aligned}
    \Delta_{\mathrm{merge}}
    &= \mathcal{M}(\Delta_1, \dots, \Delta_T), \\
    \theta_{\mathrm{merge}}
    &= \theta_0 + \Delta_{\mathrm{merge}}.
\end{aligned}
\end{equation}
CAMFT is agnostic to the choice of $\mathcal{M}$, in our experiments, we use TIES~\citep{yadav2023tiesmerging} with default hyperparameters to demonstrate that our method improves update quality independent of the merging algorithm.

\section{Experiments}
\label{sec:experiments}

\subsection{Experimental Setup}
\label{sec:setup}
\paragraph{Datasets.}
We evaluate on the TRACE benchmark~\citep{trace2023}, consisting of 7 diverse tasks:
C-STANCE~\citep{zhao-etal-2023-c},
FOMC~\citep{shah-etal-2023-trillion},
MeetingBank~\citep{hu-etal-2023-meetingbank},
ScienceQA~\citep{lu2022learn},
NumGLUE-cm and NumGLUE-ds~\citep{mishra2022numglue},
and 20Minuten~\citep{gonzales2021new}.
The statistics and evaluation metrics of these datasets are summarized in Appendix~\ref{app:datasets}.

\paragraph{Models.}
We conduct experiments on Llama-3.2-1B~\citep{llama32modelcard} as the main backbone. 
We use Qwen3-4B~\citep{qwen3} as an additional backbone and report its results in Appendix~\ref{sec:app:qwen_results}.

\paragraph{Fine-tuning methods compared.}
We compare CAMFT with two groups of baselines. 
The first group consists of representative parameter-efficient fine-tuning methods, including \textbf{LoRA}~\citep{hu2022lora}, \textbf{PiSSA}~\citep{meng2024pissa}, and \textbf{SVFT}~\citep{NEURIPS2024_48c368f1}. 
LoRA and PiSSA provide standard low-rank and spectral PEFT baselines.
SVFT is a sparse spectral baseline that updates a fixed set of SVD coordinates shared across tasks, without task-specific importance filtering or conflict-aware coordinate allocation.
The second group includes merge-friendly fine-tuning methods, including \textbf{OSRM}~\citep{zhang-zhou-2025-unraveling}, and \textbf{MergOPT}~\citep{yang2026mergopt}. 
These methods aim to improve the compatibility of independently fine-tuned task updates through regularization or merge-aware optimization.

\paragraph{Implementation details.}
All reported results are selected by validation performance from predefined hyperparameter grids, with search ranges and task-specific training schedules provided in Appendix~\ref{app:implementation}.
All methods use the same optimizer family and training protocol unless otherwise specified.
LoRA, PiSSA, and OSRM use rank 64 and have 45.1M trainable parameters per task on Llama-3.2-1B.
SVFT and CAMFT use the same target modules with sparsity $\rho=0.10$, giving 34.4M trainable parameters per task.
CAMFT searches over $\lambda$, $\gamma$, $h$, and $\alpha$ on a predefined grid (full search ranges and selected values are reported in Appendix~\ref{app:camft_hparams}), and uses 500 probing examples per task.
All experiments are run on 4 NVIDIA A100 80GB GPUs.
Unless otherwise specified, task updates are merged with TIES~\citep{yadav2023tiesmerging}, whose hyperparameters are also selected on validation data.

\subsection{Main Results}

\label{sec:main_results}

Table~\ref{tab:main_results_llama} reports a fixed-seed task-level breakdown together with three-seed aggregate results for merging seven independently trained task experts.
For the aggregate evaluation, we repeat every method with three random seeds while keeping the data splits and selected training and merging configurations fixed across runs.
CAMFT achieves the best merged macro-average score of 0.4643, outperforming MergOPT (0.4420) by 0.0223.
It also obtains the highest mergeability ratio, 83.3\% versus 79.7\% for MergOPT.
This gain does not come from simply improving single-task performance: CAMFT's single-task macro-average is 0.5573, comparable to LoRA (0.5582) and MergOPT (0.5547).
Together with the lower results of SVFT and the component ablations below, these findings support conflict-aware coordinate allocation rather than sparsity or spectral parameterization alone as the source of CAMFT's mergeability gains.

\begin{table}[t]
\centering
\small
\setlength{\tabcolsep}{6pt}
\renewcommand{\arraystretch}{1.12}
\begin{tabular}{l c c}
\toprule
\textbf{Variant} & \textbf{Merged Avg} & $\boldsymbol{\Delta}$ \\
\midrule
CAMFT (Full) & \textbf{0.464} & -- \\
\quad w/o Conflict Estimation & 0.395 & $-$0.069 \\
\quad w/o Gradient Probing & 0.376 & $-$0.088 \\
\quad w/o SVD-space Importance & 0.427 & $-$0.037 \\
\bottomrule
\end{tabular}
\caption{
Ablation study of CAMFT's core components on Llama-3.2-1B.
$\Delta$ denotes the absolute drop from full CAMFT.
}
\label{tab:ablation_core}
\end{table}

\subsection{Ablation Study}

\begin{figure}[t]
    \centering
    \includegraphics[width=0.48\textwidth]{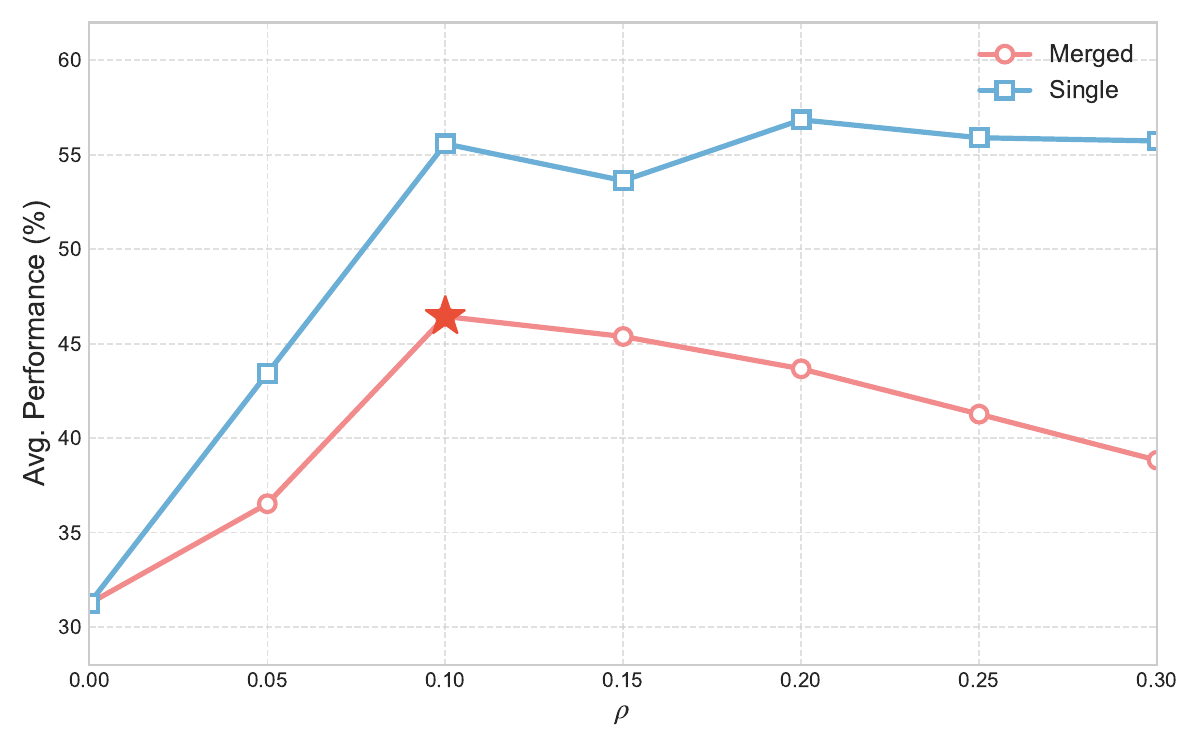}
    \caption{
Sensitivity of CAMFT to the trainable coordinate ratio $\rho$.
The star marks the \textbf{best} merged performance.
}
    \label{fig:rho_sensitivity}
\end{figure}

\paragraph{Sensitivity to trainable coordinate ratio.}
We analyze the effect of the trainable coordinate ratio $\rho$, which controls the fraction of SVD coordinates allocated to each task.
As shown in Figure~\ref{fig:rho_sensitivity}, increasing $\rho$ from $0.00$ to $0.10$ improves the merged average from $0.3125$ to $0.4644$.
Further increasing $\rho$ hurts merged performance, dropping to $0.3883$ at $\rho=0.30$, although the single-task average remains high.
This divergence suggests that larger update capacity can improve task adaptation but also reintroduce merge conflicts, supporting a compact and conflict-aware update space.

\paragraph{Core component ablation.}
We ablate three core designs of CAMFT: conflict estimation, gradient probing, and SVD-space importance estimation.
\emph{w/o Conflict Estimation} selects masks only by probing importance $\widetilde I_{t,p}$, without the conflict score $C_p$ or high-conflict exclusivity.
\emph{w/o Gradient Probing} replaces probing-based masks with random masks under the same sparsity budget.
\emph{w/o SVD-space Importance} estimates importance outside the SVD coordinate space.
All variants use the same training and TIES merging configuration as CAMFT.

As shown in Table~\ref{tab:ablation_core}, removing any component degrades merged performance.
Random masking performs worst, confirming that sparsity alone is insufficient without task-aware coordinate selection.
Importance-only masking also suffers a large drop, showing that high-gradient coordinates can still induce cross-task interference if conflict is ignored.
The degradation without SVD-space importance further suggests that the pretrained spectral basis provides a useful coordinate system for estimating and allocating task updates.
Overall, these results show that CAMFT's gains come from combining task relevance and conflict awareness in a shared SVD coordinate space.

\paragraph{Effect of probing sample size.}
We further study the effect of probing set size on CAMFT.
As shown in Table~\ref{tab:probe_samples}, very small probing sets lead to weaker performance, while a moderate number of examples is sufficient to estimate task conflicts reliably.
Using more probing samples does not further improve performance, suggesting that CAMFT does not require extensive probing.

\begin{table}[t]
\centering
\small
\setlength{\tabcolsep}{3.5pt}
\renewcommand{\arraystretch}{1.12}
\begin{tabular}{@{}l c c c c c@{}}
\toprule
\textbf{\# Samples} & \textbf{10} & \textbf{100} & \textbf{200} & \textbf{500} & \textbf{1000} \\
\midrule
\textbf{Merged Avg} & 0.4057 & 0.4276 & 0.4592 & \textbf{0.4644} & 0.4497 \\
\bottomrule
\end{tabular}
\caption{
Effect of probing sample size on merged performance.
Best result is \textbf{bolded}.
}
\label{tab:probe_samples}
\end{table}

\subsection{Robustness to merging methods.}

\label{sec:merge_robustness}

CAMFT aims to improve the intrinsic mergeability of task updates rather than relying on a particular post-hoc merging algorithm.
We therefore evaluate Full FT, MergOPT, and CAMFT under four representative merging rules: Average, Task Arithmetic, DARE, and TIES, using the same predefined hyperparameter search space for each rule.
As shown in Table~\ref{tab:merge_robustness}, CAMFT consistently outperforms both Full FT and MergOPT across all merging methods.
Its advantage under simple Average suggests that CAMFT already reduces update interference during fine-tuning, before any explicit post-hoc conflict resolution is applied.
The consistent gains under Task Arithmetic, DARE, and TIES further show that CAMFT is complementary to existing merging algorithms and improves mergeability in a method-agnostic manner.
Compared with MergOPT, CAMFT's stronger performance indicates that coordinate-level conflict-aware allocation provides benefits beyond generic merge-aware optimization.
\begin{table}[t]
\centering
\small
\setlength{\tabcolsep}{4.5pt}
\renewcommand{\arraystretch}{1.12}
\begin{tabular}{l c c c}
\toprule
\textbf{Merging Method} & \textbf{Full FT} & \textbf{MergOPT} & \textbf{CAMFT}\\
\midrule
Average & 0.368 & 0.378 & \textbf{0.389} \\
Task Arithmetic & 0.388 & 0.395 & \textbf{0.402} \\
DARE & 0.378 & 0.385 & \textbf{0.391}\\
TIES & 0.418 & 0.442 & \textbf{0.464}\\
\bottomrule
\end{tabular}
\caption{
CAMFT improves mergeability across different merging algorithms.
We report the best merged average after hyperparameter search for each method.
}
\label{tab:merge_robustness}
\end{table}
The hyperparameter search ranges for each merging method are provided in Appendix~\ref{app:merge_hparams}.

\subsection{Analysis}
\label{sec:analysis}

\begin{figure}[t]
    \centering
    \includegraphics[width=\columnwidth]{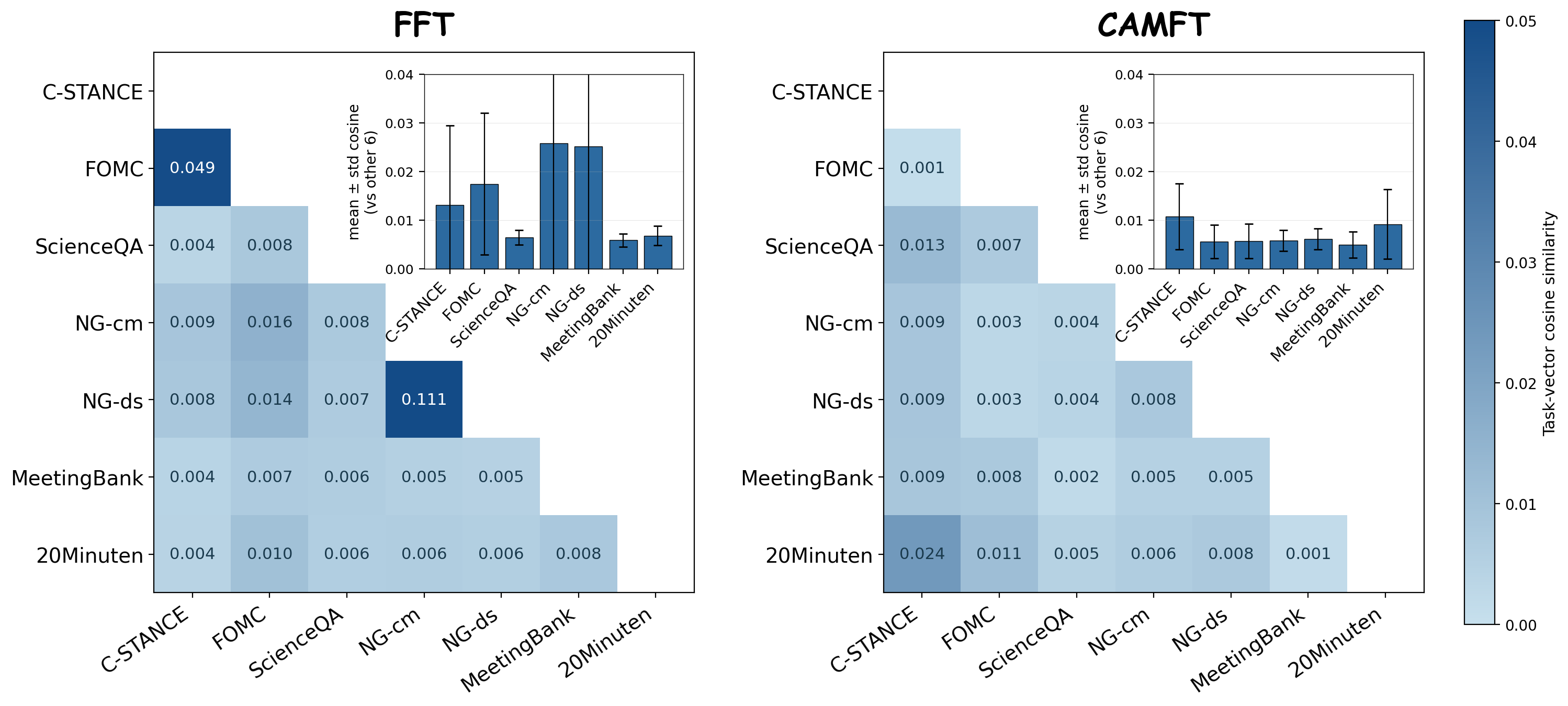}
    \vspace{2pt}
    \includegraphics[width=\columnwidth]{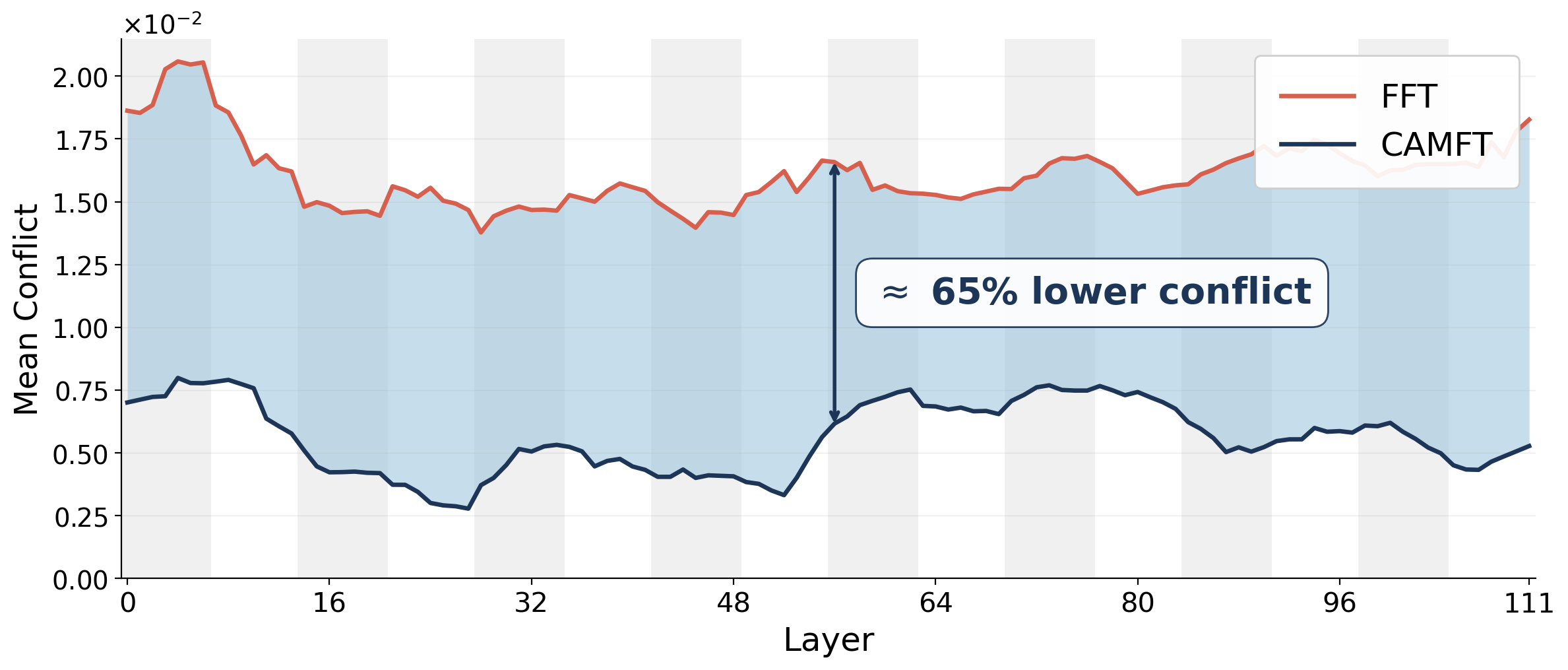}
    \caption{
    Cross-task interference before and after applying CAMFT.
    \textbf{Top:} pairwise cosine similarity of task vectors on the seven TRACE tasks under FFT (left) and CAMFT (right); 
    \textbf{Bottom:} mean off-diagonal cosine at each of the 112 linear layers of Llama-3.2-1B.
    }
\label{fig:task_conflict}
\end{figure}

\paragraph{Task-pair conflict.}
Figure~\ref{fig:task_conflict} shows that task conflicts under FFT are highly uneven and concentrated in a few task pairs. 
CAMFT markedly alleviates these concentrated conflicts, resulting in lower and more balanced per-task overlap across all seven tasks.

\paragraph{Layer-level conflict.}
The bottom panel of Figure~\ref{fig:task_conflict} shows that FFT maintains substantial cross-task interference across the network depth, whereas CAMFT consistently suppresses conflicts across layers. 
This indicates that CAMFT reduces interference broadly rather than only in a few outlier layers. 
A finer breakdown by projection type is provided in Appendix~\ref{app:per_layer_by_module}.

\section{Conclusion}
\label{sec:conclusion}

We present CAMFT, a conflict-aware fine-tuning framework that bridges parameter-efficient fine-tuning and model merging.
By constructing a shared SVD coordinate system and estimating coordinate-level task importance and conflicts before fine-tuning, CAMFT allocates sparse trainable coordinates in a merge-aware manner, preventing destructive interference before it must be repaired post hoc.
Empirical results on seven datasets show that this approach substantially improves upon existing merging strategies across multiple LLMs.

\section*{Limitations}
\label{sec:limitations}

Although CAMFT improves model merging, it has several limitations.
First, it relies on a small amount of probing data to estimate task importance and inter-task conflict, so the quality of mask allocation may depend on how representative the probing samples are.
Second, CAMFT currently uses a fixed coordinate budget across layers and tasks, while different layers and tasks may require different update capacities.
Future work could therefore explore adaptive budget allocation based on measured importance and conflict.
Finally, CAMFT assumes that all tasks are available before fine-tuning.
For incremental settings, one possible extension is to maintain a historical conflict map and allocate coordinates for each new task according to its conflict with previously learned tasks, without recomputing masks for all tasks.

As with other model merging methods, the merged model may inherit or combine undesirable behaviors from the base model or individual task experts, including biases, hallucinations, or unsafe capabilities. Since merging multiple experts can lead to behavior that is not fully predictable from each individual model, practical deployment should include task-specific safety evaluation and standard risk assessment. Our experiments are limited to academic benchmarks and do not involve high-stakes deployment.

\section*{Acknowledgments}
This work was supported by the CAS Project for Young Scientists in Basic Research (Grant No.~YSBR-083) and Lingang Laboratory (Grant No.~LGL-2616-04).

\bibliography{custom}

\clearpage

\appendix
\section{Datasets}
\label{app:datasets}

Table~\ref{tab:dataset_statistics} summarizes the statistics of the seven TRACE tasks used in our experiments, including the source domain, language, average context length, and the evaluation metric used for each task.
For all metrics in this table, larger values indicate better performance.
All task scores are normalized to $[0,1]$ before averaging in the main experiments.

\begin{table*}[t]
\centering
\small
\setlength{\tabcolsep}{6pt}
\renewcommand{\arraystretch}{1.10}
\begin{tabular}{lllll}
\toprule
\textbf{Dataset} & \textbf{Source} & \textbf{Language} & \textbf{Avg. Len.} & \textbf{Metric} \\
\midrule
\multicolumn{5}{l}{\textit{Domain-specific}} \\
ScienceQA
& Science 
& English 
& 210 
& Accuracy \\

FOMC
& Finance 
& English 
& 51 
& Accuracy \\

MeetingBank
& Meeting 
& English 
& 2,853 
& ROUGE-L \\

\midrule
\multicolumn{5}{l}{\textit{Multi-lingual}} \\
C-STANCE
& Social media 
& Chinese 
& 127 
& Accuracy \\

20Minuten
& News 
& German 
& 382 
& SARI \\

\midrule
\multicolumn{5}{l}{\textit{Mathematical reasoning}} \\
NumGLUE-cm 
& Math 
& English 
& 32 
& Accuracy \\

NumGLUE-ds
& Math 
& English 
& 21 
& Accuracy \\

\bottomrule
\end{tabular}
\caption{
An overview of dataset statistics in experiments.
\textbf{Source} indicates the origin of the context.
\textbf{Avg. Len.} denotes the average length in words for English and German datasets,
and in characters for Chinese.
\textbf{SARI} is a metric specific to simplification.
For all metrics, the larger the corresponding value, the better.
}
\label{tab:dataset_statistics}
\end{table*}

\section{Implementation Details}
\label{app:implementation}

\subsection{Training Schedules}
\label{app:training_schedule}

Table~\ref{tab:task_training_schedule} reports the task-specific training schedules.
We select the number of epochs separately for each task because the tasks differ in dataset size and convergence behavior.
For each task, we select the final configuration with the best validation performance before merging and then apply the same TIES merging protocol as in the main experiments.

\begin{table}[ht]
\centering
\small
\begin{tabular}{lc}
\toprule
Task & Epochs  \\
\midrule
C-STANCE & 5  \\
FOMC & 3  \\
ScienceQA & 7  \\
NumGLUE-cm & 5  \\
NumGLUE-ds & 5  \\
MeetingBank & 7  \\
20Minuten & 7  \\
\bottomrule
\end{tabular}
\caption{Task-specific training schedules.}
\label{tab:task_training_schedule}
\end{table}

\subsection{Trainable Parameter Counting}
\label{app:param_count}

For LoRA, PiSSA, and OSRM, we use rank 64 on all attention and MLP linear layers.
For a linear layer with input dimension $d_{\mathrm{in}}$ and output dimension $d_{\mathrm{out}}$, the number of trainable LoRA-style parameters is
\begin{equation}
    r(d_{\mathrm{in}}+d_{\mathrm{out}}).
\end{equation}
For Llama-3.2-1B, the adapted modules in each transformer block are
$q_{\mathrm{proj}}, k_{\mathrm{proj}}, v_{\mathrm{proj}}, o_{\mathrm{proj}}, gate_{\mathrm{proj}}, up_{\mathrm{proj}}$, and $down_{\mathrm{proj}}$.
Using hidden size 2048, key/value projection size 512, intermediate size 8192, 16 layers, and rank $r=64$, the trainable parameter count is
\begin{equation}
\begin{aligned}
N_{\mathrm{LoRA}}
&=
64 \times 16 \times [
(2048+2048) \\
&\quad + 2(2048+512)
+ (2048+2048) \\
&\quad + 3(2048+8192)] \\
&= 45{,}088{,}768.
\end{aligned}
\end{equation}
We therefore report LoRA, PiSSA, and OSRM as 45.1M trainable parameters per task on Llama-3.2-1B.
Full FT and MergOPT update the full model and are marked as Full.
CAMFT trains only the selected SVD-coordinate entries.
For a target linear layer, CAMFT has $\lfloor \rho \min(d_{\mathrm{in}},d_{\mathrm{out}})^2 \rfloor$ trainable SVD-coordinate entries.
Using the same target modules as LoRA/PiSSA/OSRM and $\rho=0.10$, the CAMFT parameter count is
\begin{equation}
\begin{aligned}
N_{\mathrm{CAMFT}}
&=
0.10 \times 16 \times
\left(2\cdot 512^2 + 5\cdot 2048^2\right) \\
&= 34{,}393{,}293.
\end{aligned}
\end{equation}
We therefore report CAMFT as 34.4M trainable parameters per task.

\section{Hyperparameter Search Details}
\label{app:hparams}

We tune hyperparameters for fine-tuning, merging, and CAMFT-specific components on predefined grids and select configurations based on validation performance.
The same merging-side search space is shared across all fine-tuning methods to ensure a fair comparison.
Sections~\ref{app:ft_hparams}, \ref{app:merge_hparams}, and~\ref{app:camft_hparams} report the search ranges for the three groups respectively.

\subsection{Fine-Tuning Hyperparameters}
\label{app:ft_hparams}

Table~\ref{tab:ft_hparams} reports the hyperparameter search ranges used for Full FT, LoRA, PiSSA, OSRM, and SVFT.
For each task, we select the configuration with the best validation performance before merging and then apply the same TIES merging protocol as in the main experiments.
LoRA, PiSSA, and OSRM use rank $r=64$ on all attention and MLP linear layers (45.1M trainable parameters per task on Llama-3.2-1B).

\begin{table}[ht]
\centering
\small
\setlength{\tabcolsep}{4pt}
\renewcommand{\arraystretch}{1.08}
\begin{tabular}{ll}
\toprule
\textbf{Hyperparameter} & \textbf{Search Grid} \\
\midrule
Learning rate (Full FT)
& $\{1\mathrm{e}{-5},\ 5\mathrm{e}{-5},\ 1\mathrm{e}{-4}\}$ \\
Learning rate (PEFT)
& $\{5\mathrm{e}{-5},\ 1\mathrm{e}{-4},\ 5\mathrm{e}{-4}\}$ \\
Weight decay
& $\{0,\ 0.01\}$ \\
LoRA $\alpha$ (LoRA/PiSSA/OSRM)
& $\{16,\ 32,\ 64\}$ \\
Batch size
& $16$ \\
Optimizer
& AdamW \\
\bottomrule
\end{tabular}
\caption{
Hyperparameter search grids for fine-tuning baselines.
The same grid is used across all tasks; per-task best configurations are selected on validation.
}
\label{tab:ft_hparams}
\end{table}

\subsection{Merging Hyperparameters}
\label{app:merge_hparams}

For a fair comparison, we tune the key hyperparameters of each merging algorithm on the same predefined grid for all fine-tuning methods.
For Average and Task Arithmetic, we tune the scaling coefficient.
For DARE, we tune the density of retained task-vector entries and the scaling coefficient.
For TIES, we tune the density used for trimming and the scaling coefficient.
The detailed search grids are shown in Table~\ref{tab:merge_hparams}.
For each method, we report the best merged average over validation tasks.

\begin{table}[ht]
\centering
\small
\setlength{\tabcolsep}{4pt}
\renewcommand{\arraystretch}{1.08}
\begin{tabular}{@{}lp{0.62\linewidth}@{}}
\toprule
\textbf{Merging Method} & \textbf{Hyperparameters} \\
\midrule
Average
& $-$ \\
Task Arithmetic
& $s \in \{0.2, 0.4, 0.6, 0.8, 1.0\}$ \\
DARE
& $d \in \{0.1, 0.2, 0.25, 0.3, 0.5\}$; \newline
  $s \in \{0.2, 0.4, 0.6, 0.8, 1.0\}$ \\
TIES
& $d \in \{0.1, 0.2, 0.25, 0.3, 0.5\}$; \newline
  $s \in \{0.2, 0.4, 0.6, 0.8, 1.0\}$ \\
\bottomrule
\end{tabular}
\caption{
Hyperparameter search grids for different merging algorithms.
The same search space is used for all fine-tuning methods.
}
\label{tab:merge_hparams}
\end{table}

\subsection{CAMFT-specific Hyperparameters}
\label{app:camft_hparams}

CAMFT introduces four method-specific hyperparameters: the specificity reward $\gamma$ and the conflict penalty $\lambda$ in the selection score (Eq.~\ref{eq:selection_score}), the high-conflict exclusivity ratio $h$ (Section~\ref{sec:mask_allocation}), and the update scaling factor $\alpha$ (Eq.~\ref{eq:svd_update}).
Because $\gamma$, $\lambda$, and $h$ govern \emph{cross-task} coordinate allocation, they are shared across all tasks within a single merging run; only $\alpha$, which rescales each task's reconstructed update, is tuned per task.
The total joint search budget is bounded by $|\gamma|\cdot|\lambda|\cdot|h|\cdot|\alpha|=4\cdot4\cdot3\cdot4=192$ configurations per backbone, comparable in scale to the joint (learning rate, weight decay, $\alpha$) grid used for LoRA-style baselines in Appendix~\ref{app:ft_hparams}.

Table~\ref{tab:camft_search_grid} lists the search grid, and Table~\ref{tab:camft_selected_hparams} reports the final values used in the main experiments on both backbones.
The same global $(\gamma,\lambda,h)$ configuration was selected on Llama-3.2-1B and Qwen3-4B, suggesting that the conflict-aware allocation is not sensitive to the specific backbone scale.

\begin{table}[ht]
\centering
\small
\setlength{\tabcolsep}{8pt}
\renewcommand{\arraystretch}{1.10}
\begin{tabular}{ll}
\toprule
\textbf{Hyperparameter} & \textbf{Search Grid} \\
\midrule
Specificity reward $\gamma$
& $\{0.5,\ 1.0,\ 2.0,\ 4.0\}$ \\
Conflict penalty $\lambda$
& $\{0.5,\ 1.0,\ 2.0,\ 4.0\}$ \\
Exclusivity ratio $h$ (\%)
& $\{5,\ 10,\ 20\}$ \\
Scaling factor $\alpha$
& $\{0.5,\ 1.0,\ 2.0,\ 4.0\}$ \\
\bottomrule
\end{tabular}
\caption{
Search grids for CAMFT-specific hyperparameters.
$\gamma$, $\lambda$, $h$ are shared across tasks; $\alpha$ is tuned per task.
}
\label{tab:camft_search_grid}
\end{table}

\begin{table}[ht]
\centering
\small
\setlength{\tabcolsep}{6pt}
\renewcommand{\arraystretch}{1.10}
\begin{tabular}{lcc}
\toprule
\textbf{Hyperparameter} & \textbf{Llama-3.2-1B} & \textbf{Qwen3-4B} \\
\midrule
\multicolumn{3}{l}{\textit{Global (shared across tasks)}} \\
$\gamma$ & 1.0 & 1.0 \\
$\lambda$ & 2.0 & 2.0 \\
$h$ (\%) & 10 & 10 \\
\midrule
\multicolumn{3}{l}{\textit{Per-task scaling factor $\alpha$}} \\
C-STANCE   & 1.0 & 1.0 \\
FOMC       & 1.0 & 1.0 \\
ScienceQA  & 2.0 & 2.0 \\
NumGLUE-cm & 2.0 & 2.0 \\
NumGLUE-ds & 2.0 & 1.0 \\
MeetingBank & 1.0 & 1.0 \\
20Minuten  & 1.0 & 1.0 \\
\bottomrule
\end{tabular}
\caption{
Final CAMFT hyperparameter values used in the main experiments.
Global values for $\gamma$, $\lambda$, $h$ are shared across all tasks within a merging run; the per-task $\alpha$ values are reported in the bottom block.
}
\label{tab:camft_selected_hparams}
\end{table}

\clearpage
\twocolumn[{%
\section{Additional Experimental Results}
\label{app:additional_results}

\subsection{Results on Qwen3-4B}
\label{sec:app:qwen_results}

Table~\ref{tab:qwen_results} reports results on Qwen3-4B under the same seven-task TRACE merging protocol.

\begin{center}
\begin{minipage}{\textwidth}
\centering
\small
\setlength{\tabcolsep}{4.0pt}
\renewcommand{\arraystretch}{1.10}
\begin{tabular}{lccccccccc}
\toprule
& \multicolumn{7}{c}{\textbf{Single-task Performance}} & & \\
\cmidrule(lr){2-8}
\textbf{Method}
& \textbf{C-ST}
& \textbf{FOMC}
& \textbf{SciQA}
& \textbf{NG-cm}
& \textbf{NG-ds}
& \textbf{MtgBk}
& \textbf{20Min}
& \textbf{Avg}
& \textbf{Merge.} \\
\midrule
Full FT
& 0.536 & 0.712 & 0.842 & 0.491 & 0.651 & 0.684 & 0.439 & \textbf{0.622} & -- \\
LoRA
& 0.512 & \textbf{0.724} & \textbf{0.881} & 0.362 & 0.663 & 0.702 & 0.431 & 0.611 & -- \\
PiSSA
& 0.501 & 0.704 & 0.868 & 0.381 & 0.647 & \textbf{0.711} & 0.417 & 0.604 & -- \\
SVFT
& 0.477 & 0.681 & 0.691 & 0.351 & 0.593 & 0.583 & 0.399 & 0.539 & -- \\
OSRM$^\dagger$
& 0.505 & 0.718 & 0.873 & 0.345 & 0.626 & 0.681 & 0.422 & 0.596 & -- \\
MergOPT$^\dagger$
& 0.532 & 0.719 & 0.751 & 0.497 & 0.646 & 0.693 & 0.429 & 0.610 & -- \\
CAMFT$^\dagger$
& \textbf{0.542} & 0.721 & 0.738 & \textbf{0.548} & \textbf{0.682} & 0.596 & \textbf{0.438} & 0.609 & -- \\
\midrule
& \multicolumn{7}{c}{\textbf{Merged Performance}} & & \\
\cmidrule(lr){2-8}
\textbf{Method}
& \textbf{C-ST}
& \textbf{FOMC}
& \textbf{SciQA}
& \textbf{NG-cm}
& \textbf{NG-ds}
& \textbf{MtgBk}
& \textbf{20Min}
& \textbf{Avg}
& \textbf{Merge.} \\
\midrule
Full FT
& 0.492 & 0.647 & 0.583 & 0.371 & 0.517 & 0.377 & 0.421 & 0.487 & 78.3\% \\
LoRA
& 0.487 & 0.665 & 0.244 & 0.163 & 0.612 & 0.092 & 0.424 & 0.384 & 62.9\% \\
PiSSA
& 0.486 & 0.657 & 0.331 & 0.252 & 0.588 & 0.246 & 0.414 & 0.425 & 70.3\% \\
SVFT
& 0.443 & 0.630 & 0.414 & 0.271 & 0.405 & 0.317 & 0.392 & 0.410 & 76.1\% \\
OSRM$^\dagger$
& 0.476 & \textbf{0.699} & 0.356 & \textbf{0.384} & 0.579 & 0.332 & 0.424 & 0.464 & 77.9\% \\
MergOPT$^\dagger$
& 0.514 & 0.648 & 0.493 & 0.317 & \textbf{0.642} & 0.516 & 0.423 & 0.508 & 83.3\% \\
CAMFT$^\dagger$
& \textbf{0.523} & 0.662 & \textbf{0.557} & 0.365 & 0.637 & \textbf{0.610} & \textbf{0.429} & \textbf{0.540} & \textbf{88.7\%} \\
\bottomrule
\end{tabular}
\captionsetup{hypcap=false}
\captionof{table}{
Performance comparison on TRACE 7-task benchmark with Qwen3-4B.
\textbf{Single}: single-task performance before merging.
\textbf{Merged}: performance after merging all 7 tasks with TIES.
\textbf{Merge.}: mergeability (Merged Avg / Single Avg $\times$ 100\%).
All task metrics are normalized to [0,1].
Best results are \textbf{bolded}.
$^\dagger$: methods designed for mergeability.
}
\label{tab:qwen_results}
\end{minipage}
\end{center}
}]

\subsection{Per-layer Conflict by Projection Type}
\label{app:per_layer_by_module}

\begin{figure*}[t]
    \centering
    \includegraphics[width=\textwidth]{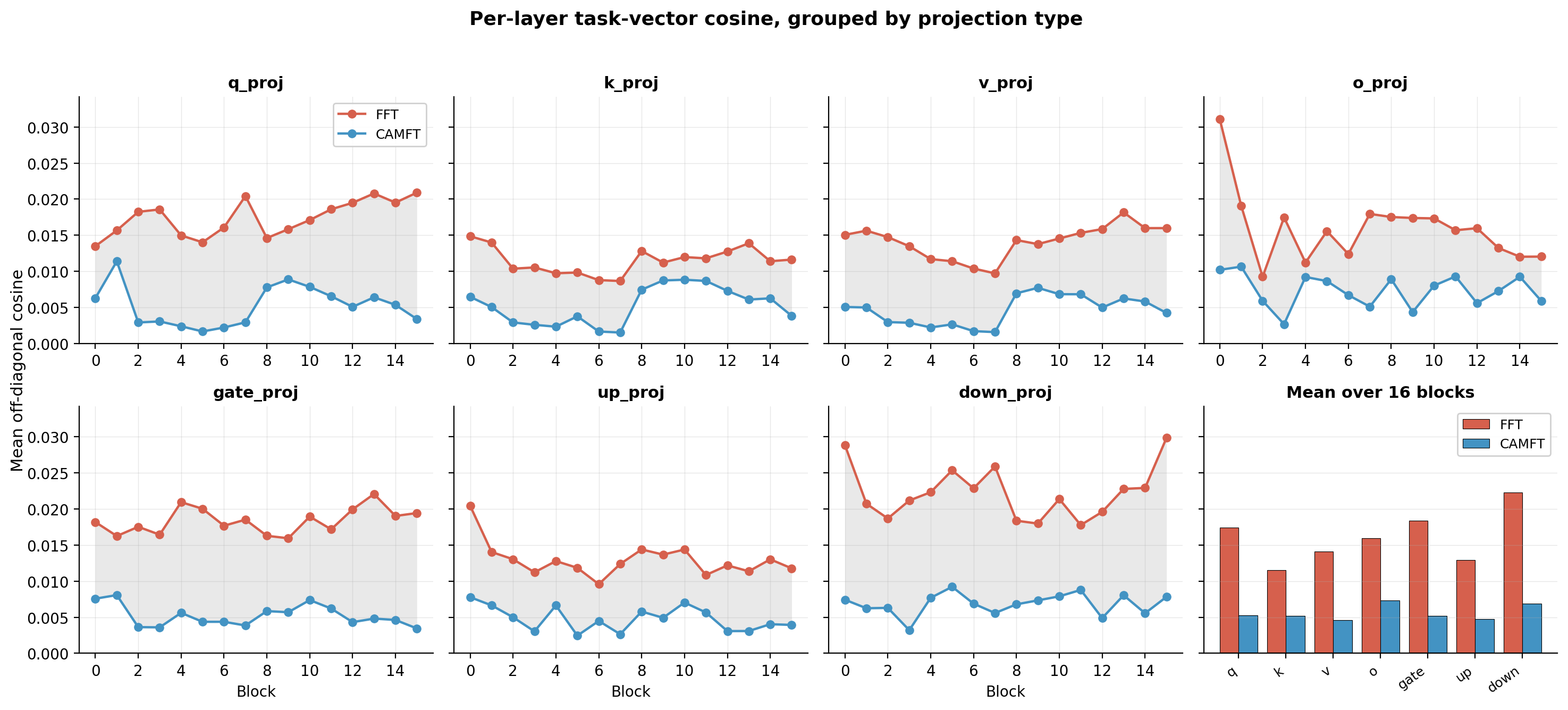}
    \caption{
    Per-layer task-vector cosine, decomposed by projection type. Each of the seven panels plots the 16 transformer blocks for one projection; the bottom-right panel reports the per-projection mean over all blocks.
    }
    \label{fig:per_layer_by_module}
\end{figure*}

Figure~\ref{fig:per_layer_by_module} decomposes the per-layer conflict in the bottom panel of Figure~\ref{fig:task_conflict} along the seven projection types (q, k, v, o, gate, up, down).
The reduction from FFT to CAMFT is consistent across all projection types and all 16 transformer blocks, with the largest absolute gap observed on \texttt{q\_proj} and \texttt{down\_proj}.
The bottom-right panel summarises the per-projection mean over blocks.

\FloatBarrier
\end{document}